\documentclass[twocolumn,10pt]{asme2e}

\confshortname{ISFA 2020}
\conffullname{International Symposium on Flexible Automation}
\confdate{5-9}
\confmonth{July}
\confyear{2020}
\confcity{Chicago}
\confcountry{USA}
\papernum{ISFA2020-9657}
\title{A Real-Time Receding Horizon Sequence Planner for Disassembly in A Human-Robot Collaboration Setting}
\usepackage{graphicx}
\usepackage{amsmath} 
\usepackage{graphicx}
\usepackage{float}
\usepackage{algpseudocode}
\usepackage{algorithm}
\usepackage{xcolor}
\usepackage{array}
\usepackage{ulem}
\usepackage{comment}
\usepackage{booktabs}
\usepackage{epstopdf}
\usepackage{url}

\author{Meng-Lun Lee
    \affiliation{
	Department of Mechanical and\\
	Aerospace Engineering\\
	University at Buffalo\\
	Buffalo, NY 14260\\
    Email: menglunl@buffalo.edu
    }	
}

\author{Sara Behdad
    \affiliation{
	Department of Environmental\\ Engineering Sciences\\
	University of Florida\\
	Gainesville, FL 32611\\
    Email: sarabehdad@ufl.edu
    }	
}

\author{Xiao Liang
    \affiliation{
	Department of Civil, Structural and\\
	Environmental Engineering\\
	University at Buffalo\\
	Buffalo, NY 14260\\
    Email: liangx@buffalo.edu
    }	
}
\author{Minghui Zheng\thanks{Address all correspondence to this author.}
    \affiliation{
	Department of Mechanical and\\
	Aerospace Engineering\\
	University at Buffalo\\
	Buffalo, NY 14260\\
    Email: mhzheng@buffalo.edu
    }	
}

\begin{document}

\maketitle    
\graphicspath{ {./figs/} }
\begin{abstract}
{\it Product disassembly is a labor-intensive process and is far from being automated. Typically, disassembly is not robust enough to handle product varieties from different shapes, models, and physical uncertainties due to component imperfections, damage throughout component usage, or insufficient product information. To overcome these difficulties and to automate the disassembly procedure through human-robot collaboration without excessive computational cost, this paper proposes a real-time receding horizon sequence planner that distributes tasks between robot and human operator while taking real-time human motion into consideration. The sequence planner aims to address several issues in the disassembly line, such as varying orientations, safety constraints of human operators, uncertainty of human operation, and the computational cost of large number of disassembly tasks. The proposed disassembly sequence planner identifies both the positions and orientations of the to-be-disassembled items, as well as the locations of human operator, and obtains an optimal disassembly sequence that follows disassembly rules and safety constraints for human operation. Experimental tests have been conducted to validate the proposed planner: the robot can locate and disassemble the components following the optimal sequence, and consider explicitly human operator's real-time motion, and collaborate with the human operator without violating safety constraints.}
\end{abstract}

\vspace{-10pt}
\section*{INTRODUCTION}

Environmentally conscious manufacturing (ECM) \cite{gungor:99} and product recycling over the past decade have become an obligation for many manufacturers to follow government regulations \cite{Torres:09}. And enterprises increase attention to recycle end of life (EOL) products \cite{pullen:05} by consumers and develop cost-efficient ways for recycling and/or recovering them. Therefore, the efficiency and cost-effectiveness play a crucial role in both product disassembly and reassembly processes.

Assembly/disassembly sequences consist of actions with ordering. An action is usually determined from an engineering point of view, such as the manufacturing of a product from sub-assemblies into one complete unit, or the separation of finished goods into sub-assemblies \cite{lambert:03}. Multiple feasible assembly/disassembly sequences may exist in an ordinary product, like the hard disk and the CD player. The number of sequence grows larger as the complexity of the product increases. Because it is difficult to represent every sequence on an individual basis, a systematic and efficient approach to illustrate all possible sequences and to achieve the optimal one is required. Subsequently, many algorithms like incidence matrix \cite{fulkerson:65} \cite{Lue:08}, AND/OR graphs \cite{de:90}, directed graphs \cite{Gansner:93}, graph visualization \cite{berg2013disassembly}, mixed-integer nonlinear programming \cite{behdad2014leveraging}, and precedence relationships \cite{lambert:02} \cite{moore:01} have been proposed. Several studies also use assembly drawings from computer-aided design models \cite{lin:93} for task planning. For instance, to find the optimal disassembly sequence, many researches are carried out based on the modified Traveling Salesman Problem \cite{navtn:94}, which explores the sequence of a salesperson traveling from one city to other cities without passing the same city twice. Most of the studies aim to automatically find the optimal sequence based on information that is not unknown, such as a fixed initial task, several common/parallel tasks, and the rules for the assembly/disassembly tasks \cite{Torres:09}. However, these studies take no consideration for the human-robot collaboration or the safety constraints. \vspace{-10pt}

\begin{figure}[!htbp]
	\centering
	\includegraphics[width=1.5in]{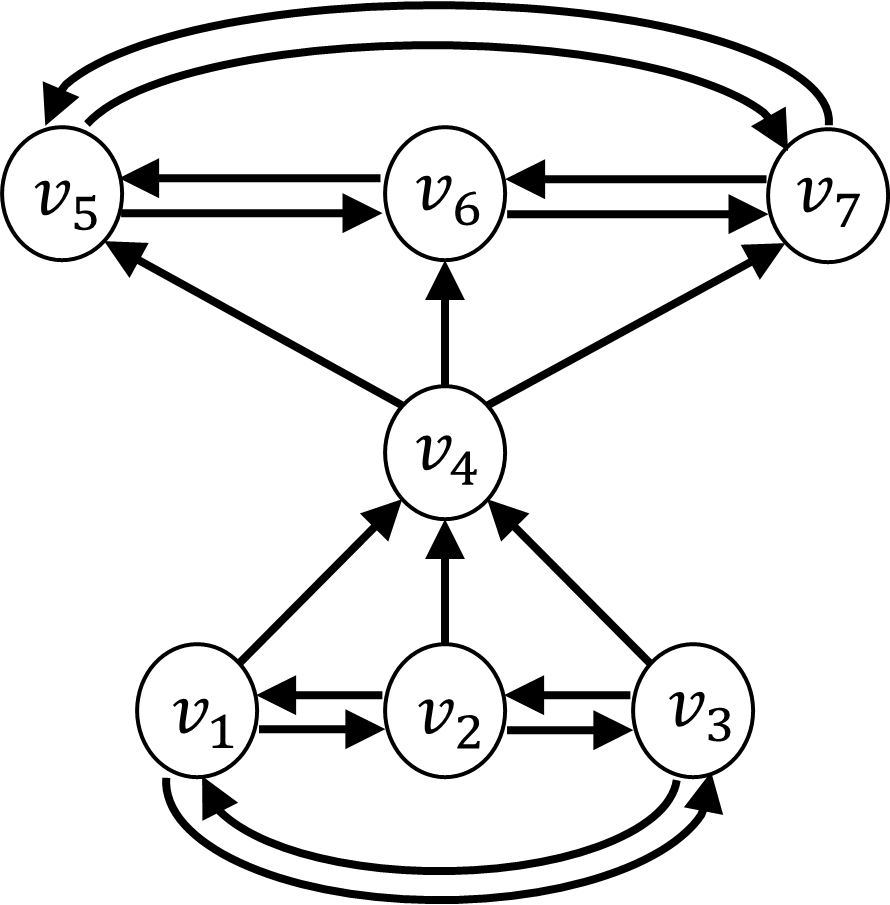}
	\caption{DIRECTED GRAPH}
	\vspace{-15pt}
	\label{p1}
\end{figure}

Typically, the assembly method is unique \cite{papa:14}: the assembly line is conducted by a work-cell format with serial workstations and distinctive fixtures \cite{de:87} such that the orientation of all the disassembling objects will be consistent. This is usually processed by robots that are programmed for repetitive tasks \cite{wang2018robust,wang2016robust}. The disassembly process, on the contrary, is more challenging for many reasons: (1) a range of used products (e.g., electronic wastes) may be delivered to the same disassembly line to minimize the cost of building separate disassembly lines; (2) disassembly is sometimes carried out incompletely to collect valuable parts \cite{lambert:03}; likewise, (3) the products to be recycled may have components with dangerous substances and need to be handled cautiously; (4) different products may be imported to the same disassembly line with inconsistent orientations; (5) a robot may take unnecessary or longer trips to execute the designated disassembly sequence, regardless of other existing disassembly sequences that may have a shorter path or working time. Thus, the majority of the disassembly lines are operated by human operators \cite{soh:14} rather than by automated systems.

Additionally, if there is more than one starting point for a disassembly, more than one hierarchical graph model \cite{Puente:01} may be represented for the precedence relation of disassembly tasks. For example, given seven tasks: $v_1$, $v_2$, $\hdots$, $v_7$, we assume there are two disassembly rules applied:
\begin{itemize}
	\item Rule 1: Remove $v_1$ or $v_2$ or $v_3$
	$\rightarrow$ Remove $v_4$
	\item Rule 2: Remove $v_4$ 
	$\rightarrow$ Remove $v_5$ or $v_6$ or $v_7$
\end{itemize}
where Rule 1 states that task $v_1$, $v_2$ and $v_3$ must be executed before task $v_4$; Likewise, Rule 2 shows that task $v_4$ must proceed prior to task $v_5$, $v_6$ and $v_7$. The associated directed-graph is shown in Figure \ref{p1}, where the arcs are the disassembly tasks and the arrows represent the precedence relations among the seven tasks. The tasks can be defined as ``parallel tasks" or ``common tasks" \cite{Torres:09}; in this example, $v_1$, $v_2$ and $v_3$ in Rule 1 are ``parallel tasks" and the $v_5$, $v_6$ and $v_7$ in Rule 2 are also ``parallel tasks" while $v_4$ in both Rule 1 and Rule 2 is a ``common task". 

\begin{figure}[!htbp]
	\centering
	\includegraphics[width=3.4in]{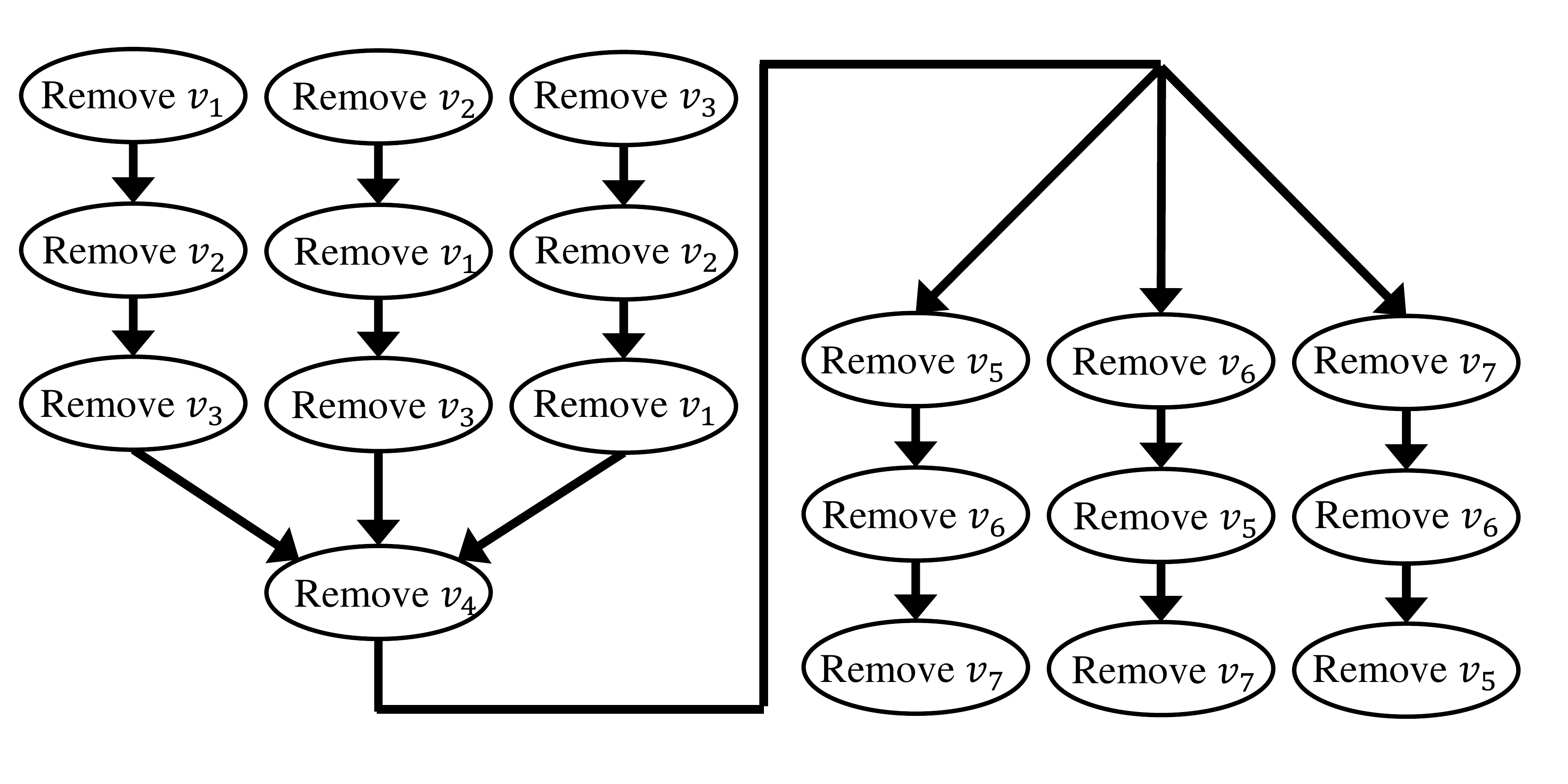}
	\caption{POSSIBLE SEQUENCES OF SEVEN TASKS}
	\label{fig:s2}
	\vspace{-15pt}
\end{figure}

Because either $v_1$, $v_2$ or $v_3$ can be chosen as the beginning task and $v_5$, $v_6$ or $v_7$ will become the final task, the graph model is represented as Figure \ref{fig:s2}. The task sequence will not be unique because there are many possible starting and ending points. 

\vspace{-15pt}

In addition, it is assumed that the products with varied orientation would be fed into the disassembly line. Therefore, a feasible sequence with multiple starting points and ending points implies the existence of options to improve disassembly task efficiency. The working time at each disassembly task affects the cost of tasks \cite{Torres:09}. For instance, the actions of targeting, grabbing, and releasing a screw could take different lengths of time based on whether they are performed by a robot or by a human. Apart from the working time, other factors such as component geometric complexity should also be considered. If the product requires a diverse set of orientations and safe conditions, it is evident that both complexity of task execution and travel time between each disassembly may not remain constant. Moreover, human operation may be prohibited due to either unsafe work environment \cite{lambert:02} or potentially hazardous materials. Thus, a decision-maker capable of considering safety constraints and assigning disassembly tasks between human and robot workers should be developed.

A conventional product will have an initially high computational cost if it has a considerable number of disassembly tasks. Unfortunately, considering optimization for a given optimal disassembly sequence solution may often increase this computational cost significantly. To overcome this issue, we adapt the receding horizon control technique such that it can not only solve the optimal sequence locally with manageable computational costs, but also consider real-time human motion when executing the first task from the local optimal sequence. The receding horizon control functions follow three main steps. Assuming that there are $T$ total disassembly tasks and the receding horizon is set as $t$ where $t \leq T$ and $t$, $T$ $\in \mathbf{Z}^+$, (1) the sequence planner generates the $t$ possible sequences at each step, (2) the local optimal sequence with minimum cost is obtained, and (3) the first sequence of the local optimal sequence determined by the sequence planner is executed by either the robot or the human operator. Additionally, an example of using the receding horizon method to reduce the computational cost can be outlined. Assuming that all the tasks are parallel tasks without using receding horizon method, the number of feasible sequences will be $T! \times 2^T$, where $T!$ = $(T) \times (T-1) \times (T-2) \times ... \times (2)\times (1)$. The $T!$ term is generated by the number of parallel tasks and $2^T$ term is caused by the collaboration of two workers: the human operator and the robot. By contrast, if the technique of receding horizon is used, the number of feasible sequences at each step will become $t! \times 2^T$. In this case, the optimal sequence search tree will be smaller because $t$ is less than $T$, contributing to computational cost reduction.

We summarize here the main contributions of this paper. This paper formulates the disassembly sequence planning into an optimization problem and solves it online in a receding horizon way via considering the uncertainty of human operation. The formulation explicitly considers various product orientations, geometric disassembly rules, safety constraints of the human operator, the quantification of the cost of human operation and robot operation, and targeting to optimize the distribution of the disassembly task between human and robot in a human-robot collaborative setting.

The remaining of this paper is structured as follows: Section 2 introduces disassembly progress considerations, including formulation of disassembly rules, conduction cost by robot and/or human operations, and consideration of safety constraints for human worker. Section 3 describes the proposed sequence planning. Section 4 presents experimental verification to validate the proposed disassembly planning. Section 5 concludes the paper.

\vspace{-10pt}
\section*{DISASSEMBLY PROCESS CONSIDERATIONS}
\begin{figure}[ht]
	\centering
	\includegraphics[width=0.52\linewidth]{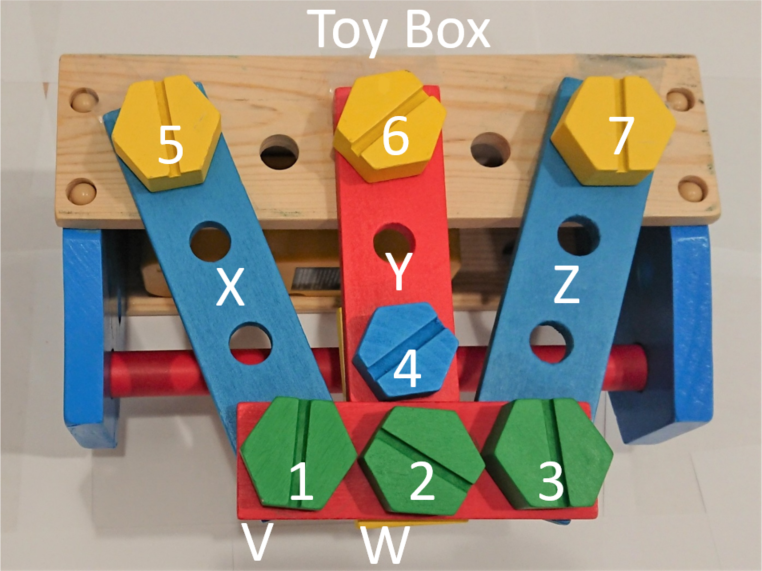}
	\includegraphics[width=0.27\linewidth]{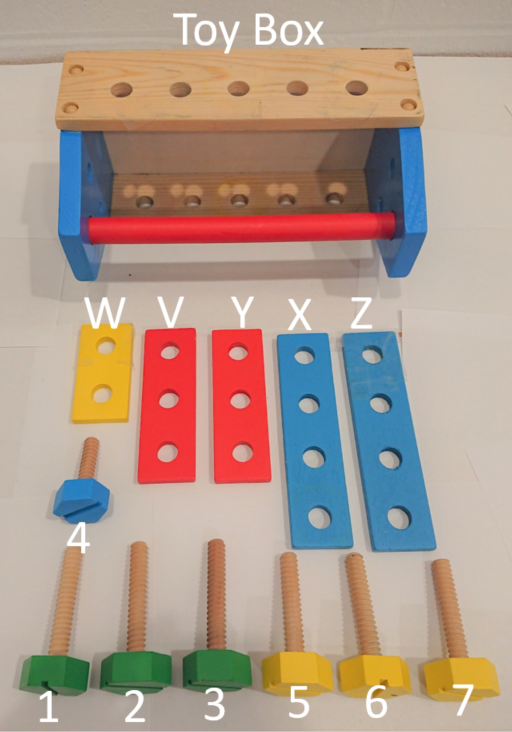}
	\caption{TO-BE-DISASSEMBLED WOODEN TOY BOX AND ITS COMPONENTS}
	\vspace{-15pt}
	\label{fig:2_1}
\end{figure}

In this paper, a wooden toy box shown in Fig. \ref{fig:2_1} is used as the to-be-disassembled object using human-robot collaboration. Before generating the disassembly sequences of the wooden box, acquiring the information of the to-be-disassembled product is essential. This includes the conditions and coordinates of its components. If this data is not captured correctly, the robot can not locate the parts to-be-disassembled and unsafe tasks may be accidentally allocated to the human operator. \vspace{-10pt}

Even though we state that it is possible to proceed with the disassembly sequence planning for distinctive products in the same disassembly line, the scope is narrowed down to a wooden toy box with known components in order to highlight the optimal sequence planning and its validation. In this paper it is also assumed that the identification of the locations of these components on the wooden toy box is sufficient for planning the disassembly task sequence.

\vspace{-15pt}
\subsection*{Disassembly rules}
To understand the precedence constraints among each component, we have made the disassembly sample shown in Fig. \ref{fig:2_1} and the corresponding graph model is conducted in Fig. \ref{p1}. The properties of the graph model have been briefly introduced in the previous chapter. We further define the operator $R$ to be the action of disassembling a component, and the notations of $\wedge$ and $\rightarrow$ mean that the tasks on both sides of $\wedge$ must be completed before executing the task on the right hand side of $\rightarrow$. The initial rules for disassembling the toy box are (1) $R_1 \wedge R_2 \wedge R_3$ $\rightarrow$ $R_V$, (2) $R_V \wedge R_4 \rightarrow$ $R_W$, (3) $R_V \wedge R_5$ $\rightarrow$ $R_X$, (4) $R_W \wedge R_6$ $\rightarrow$ $R_Y$, (5) $R_V \wedge R_7$ $\rightarrow$ $R_Z$. Screw 1, Screw 2 and Screw 3 are the first objects to be dismantled because of no other components stacking on these screws. Similarly, Screw 5, Screw 6 and Screw 7 are the last ones to be unloaded because they support the weight of all the other screws and rectangular parts. Furthermore, we will take away the rectangular parts V, W, X, Y, and Z right after the corresponding screws being removed, so the disassembly rules can be simplified as:
\begin{itemize}
	\item Rule 1: $R_1 \wedge R_2 \wedge R_3$ $\rightarrow$ $R_4$
	\item Rule 2: $R_4$ $\rightarrow$ $R_5 \wedge R_6 \wedge R_7$
\end{itemize}
Rule 1 shows that Screw 1, Screw 2 and Screw 3 must all be  removed in advance of Screw 4. Similarly, Rule 2 means Screw 5, Screw 6 or Screw 7 can be disassembled after the removal of Screw 4.

\vspace{-10pt}
\subsection*{Cost of operations by robot and human}
To find the optimal sequence of disassembling a product, we need to define the cost for each disassembly action. It is reasonable to parameterize the cost as the combination of the traveling distance between the tasks and the robot/human operator, as well as the geometric complexity of the product. While the traveling distance between each disassembly task and the robot/human operator is measurable, the determination of geometric complexity requires some preliminary tests.

\vspace{-10pt}
\subsection*{Consideration of human operator safety}
We assume that the removal of hazardous materials may be required during the disassembly sequence. In this instance, the task planner should assign the robot to disassemble the component instead of the human operator to avoid potential injuries. Safety requirements should be taken into account regardless of the potentially higher disassembly cost generated by the robot. 

\vspace{-10pt}
\subsection*{Consideration of computational cost} 
In a previous study \cite{menglunl:19}, an optimal disassembly planning is formulated to minimize the total disassembly cost from the starting node to the final node. This method is practical only if the number of disassembly tasks is relatively small. As the number of disassembly tasks rises, the task planner may suffer from taking a significant amount of time to search for the task sequence with minimum cost. Therefore, it is necessary to determine the optimal feature sequence without explicitly evaluating all the possible uniquely ordered task sequences.

\vspace{-10pt}
\section*{DISASSEMBLY SEQUENCE PLANNING}
\begin{figure}[b]
	\vspace{-20pt}
	\centering
	\includegraphics[width=0.25\textwidth]{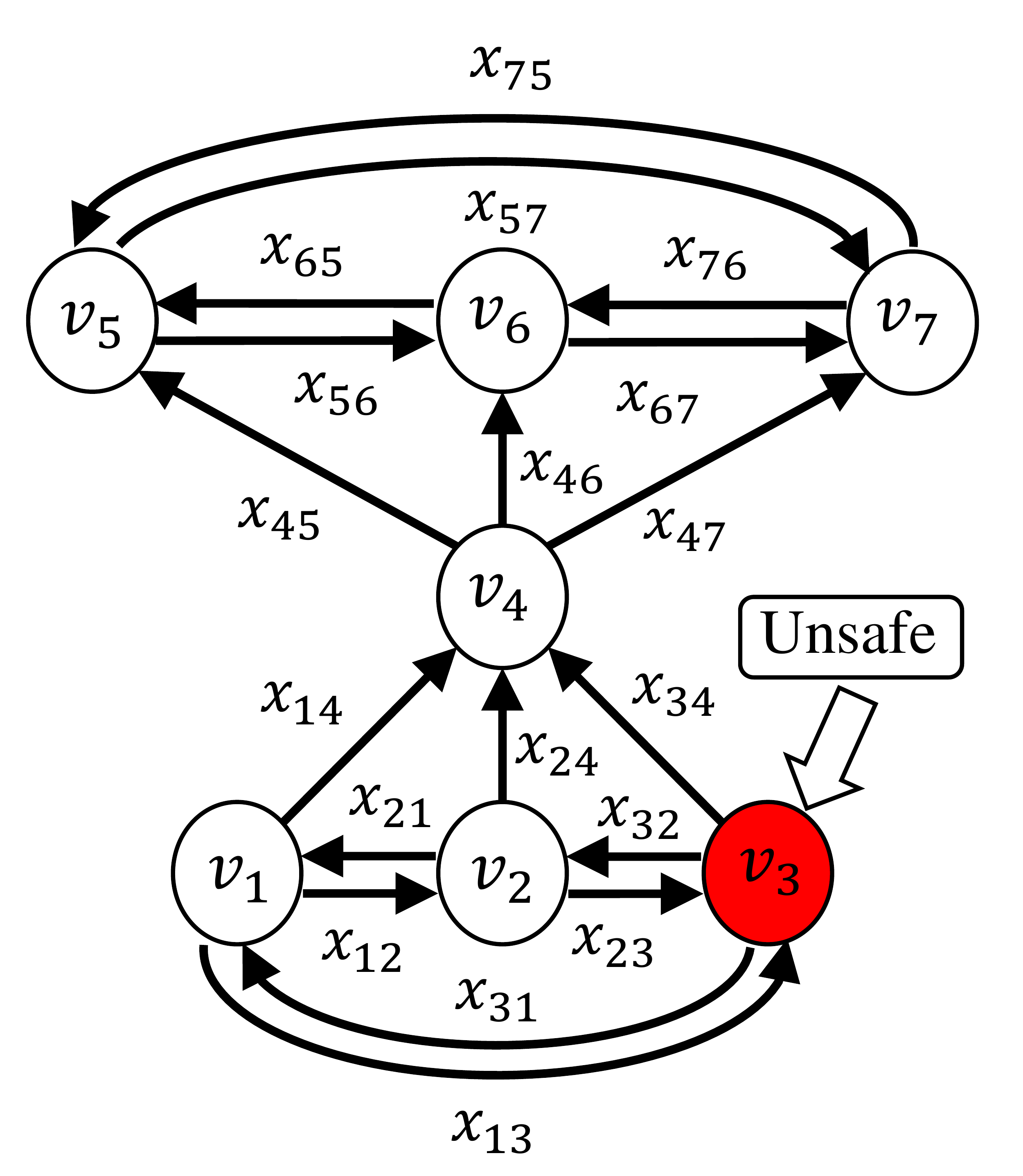}
	\caption{DISASSEMBLY SEQUENCE ILLUSTRATION}
	\label{fig:unsafe}
\end{figure}

This section first formulates the disassembly sequence planning problem into an optimization problem considering human-robot collaboration and real-time human behavior. The disassembly cost, that is assumed to be the summation of travel distances between disassembly tasks and the effort spent at each disassembly task (i.e., time and complexity) by the human operator or the robot, is minimized. 
We first introduce the parameters, operators, states, decision variables of the disassembly sequence planner, followed by the objective function below.

\vspace{3pt}
\noindent \textbf{Parameters}:\\
$T$: The number of total disassembly tasks, $T \in \mathbf{Z}^+$\\
$t$: The current step of performing a disassembly task, $t \leq T$\\
$v_t$: The $t$-th disassembly task (as shown in Fig.~\ref{fig:unsafe})\\
$N$: The number of tasks for prediction $N \leq T$\\
$c_{ij}$: Disassembly cost of task $v_j$ after $v_i$ is completed\\
$\tau$: Weighting factor in disassembly cost quantification \\

\noindent \textbf{States}:\\ 
$P(v_j)$: Location of disassembly task $v_j$\\
$q^{h}_i$: Position of human operator after $v_i$ is completed\\
$q^{r}_i$: Position of robot after $v_i$ is completed\\

\noindent \textbf{Operators}:\\
$h_{ij}$: Human's traveling cost to $v_j$ after finishing $v_i$\\
$r_{ij}$: Robot's traveling cost to $v_j$ after finishing $v_i$\\
$S^h(v_n)$: Human's labor effort spent on $v_n$ \\
$S^r(v_n)$: Robot's labor effort spent on  $v_n$ \\

\noindent \textbf{Decision Variables}: \vspace{5pt}\\
$x_{ij}=
\begin{cases}
1 \quad \text{if task from $v_i$ to $v_j$ is conducted}\\ 
0 \quad \text{otherwise}\\
\end{cases}\vspace{3pt}$ (see Fig.~\ref{fig:unsafe})\\
$\alpha_{ij}=
\begin{cases}
1 \quad \text{if task from $v_i$ to $v_j$ is conducted by robot}\\
0 \quad \text{if task from $v_i$ to $v_j$ is conducted by human}\\
\end{cases}\vspace{6pt}$\\

\noindent \textbf{Cost function}:
\vspace{-10pt}
\begin{equation} \label{eq:cost_func}
\min_{x_{ij}, \alpha_{ij}, i \neq j}  \sum_{j=0}^{T} \sum_{i=0}^{T} c_{ij} x_{ij}
\end{equation}

\noindent \textbf{Constraints}:
\vspace{-10pt}
\begin{equation} \label{eq:a}
\sum_{i=0}^{T}\left[ \sum_{j=0}^{T} x_{ij} - \sum_{k=0}^{T} x_{ki}\right] = 0 
\end{equation}
\vspace{-30pt}
\begin{equation} \label{eq:b}
x_{ij} \in \{ 0, 1 \}
\end{equation}
\vspace{-30pt}
\begin{equation} \label{eq:c}
\sum_{i=0}^{T} \sum_{j=0}^{T} x_{ij} = T
\end{equation}
\vspace{-30pt}
\begin{equation} \label{eq:d}
c_{ij} = \alpha_{ij} r_{ij} + (1-\alpha_{ij}) h_{ij}
\end{equation}
\vspace{-30pt}
\begin{equation} \label{eq:e}
\alpha_{ij} \in \{ 0, 1 \}
\end{equation}
\vspace{-30pt}
\begin{equation} \label{eq:f}
\alpha_{ij} = 1 \quad \text{if task } j \text{ is unsafe for human operation}
\end{equation}
\vspace{-30pt}
\begin{equation} \label{eq:g}
h_{ij} = \big|q^{h}_i-P(v_j)\big|+\tau S^h(v_j)
\end{equation}
\vspace{-30pt}
\begin{equation} \label{eq:h}
r_{ij} = \big|q^{r}_i-P(v_j)\big|+\tau S^r(v_j)
\end{equation}
\vspace{-20pt}

As we briefly introduced earlier, $x_{ij}$ determines whether Screw $j$ will be disassembled  right after disassembling Screw $i$, where $i,j=1,2,..,T$. In constraint (\ref{eq:a}) the terms inside the square bracket represent two constraints: (a) the number of workflow into a task is equal to the number of workflow out of the same task except for the starting task and final task of a disassembly sequence; (b) the number of workflow out of the starting task is equal to that of the workflow into the final task. Constraints (\ref{eq:b}) and (\ref{eq:c}) ensure that each disassembly is executed for only one time and all the tasks would be conducted.

Next, we consider the human-robot collaboration setting by adding constraint (\ref{eq:d}). Here, the disassembly cost $c_{ij}$ is defined by both the robot's cost $r_{ij}$ and the human operator's cost $h_{ij}$, and each of them is presumably defined by two terms as expressed in (\ref{eq:g}) and (\ref{eq:h}): (a) the distance between task $j$ and the robot/human operator after leaving task $i$ respectively; (b) the effort spent on removing screw $j$ by robot/human operator, respectively. Moreover, the decision variable $\alpha_{ij}$ in (\ref{eq:d}) and (\ref{eq:e}) are used to determine whether the task will be done by robot or by a human operator. If an unsafe condition is detected at task $v_j$, the corresponding decision variable $\alpha_{ij}$ will trigger constraint (\ref{eq:f}), so that the robot will be forced to execute the unsafe task $v_j$. For example, task of $v_3$ in Fig. \ref{fig:unsafe} is labeled as unsafe for human operation, the sequence planner in this situation will set the decision variables $\alpha_{23}$ and $\alpha_{13}$ to be 1, forcing Screw 3 to be disassembled by the robot instead of the human operator. Lastly, the $\tau$ denotes the weighting factor to balance the two terms in (\ref{eq:g}) and (\ref{eq:h}).

\begin{figure}[!htbp]
	\centering
	\includegraphics[width=0.33\textwidth]{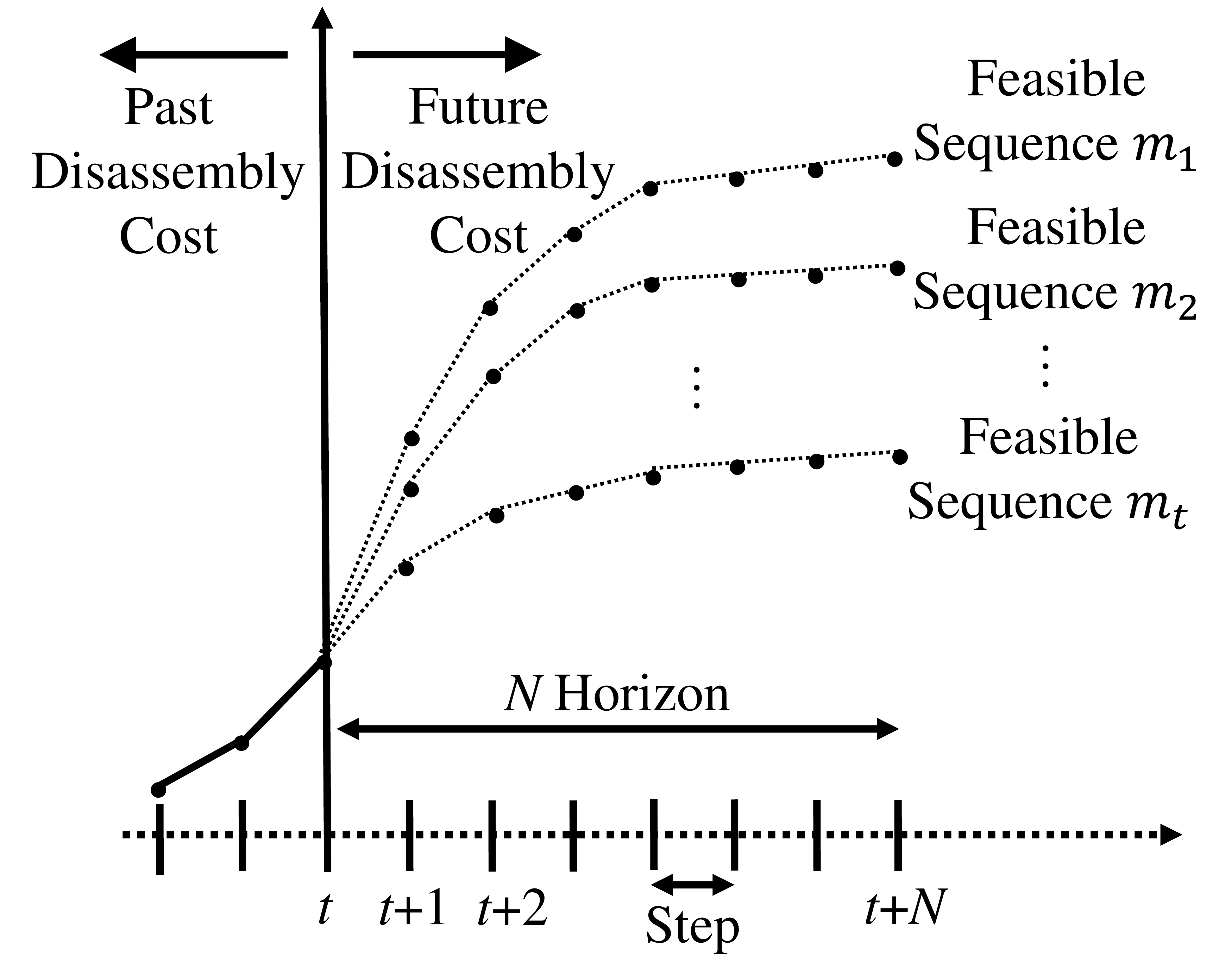}
	\caption{RECEDING HORIZON OF SEQUENCE PLANNING}	\vspace{-15pt}
	\label{fig:recede}
\end{figure}

\begin{figure}[!htbp]
	\centering
	\includegraphics[width=2.4in]{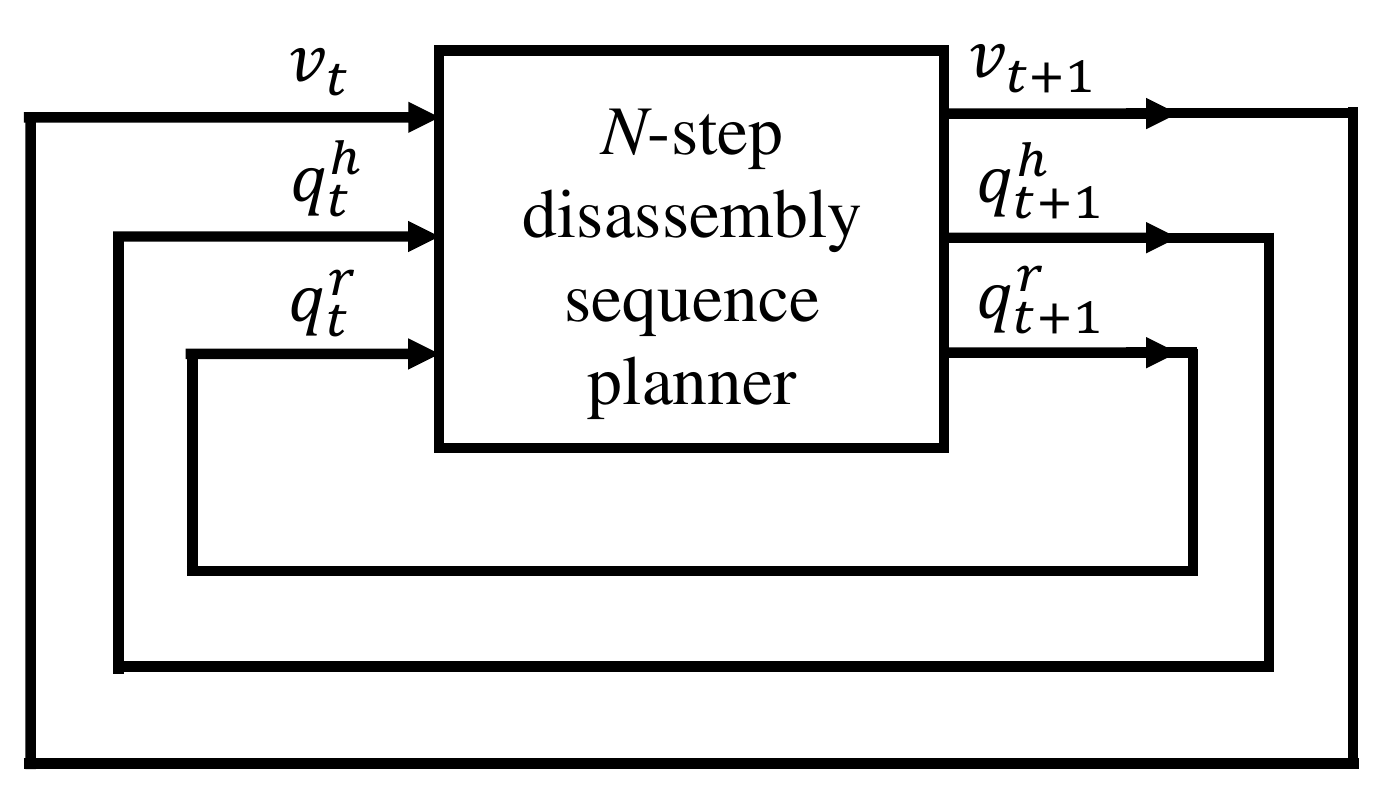}
	\caption{$N$-STEP DISASSEMBLY SEQUENCE PLANNER}\vspace{-15pt}
	\label{fig:control1}
\end{figure}

We aim to solve the above optimization problem to achieve the values of the decision variables $x_{ij}$ and $\alpha_{ij}$, such that the optimal sequence is obtained. However, directly solving the optimization is challenging especially when $T$ is large. Moreover, since the human position is changing in real-time, which may affect the optimal sequence obtained at the very beginning of the disassembly. Therefore, here we  borrow the concept of model predictive control (MPC) \cite{allgower2012} and solve this optimization problem for a shorter time horizon, denoted as $N$, at every step, and implement only the first-step solution. MPC in control and planning domain consists of an iterative, finite-horizon optimization of a plant model. 
\begin{figure}[!htbp]
	\vspace{-20pt}
	\centering
	\includegraphics[width=0.68\linewidth]{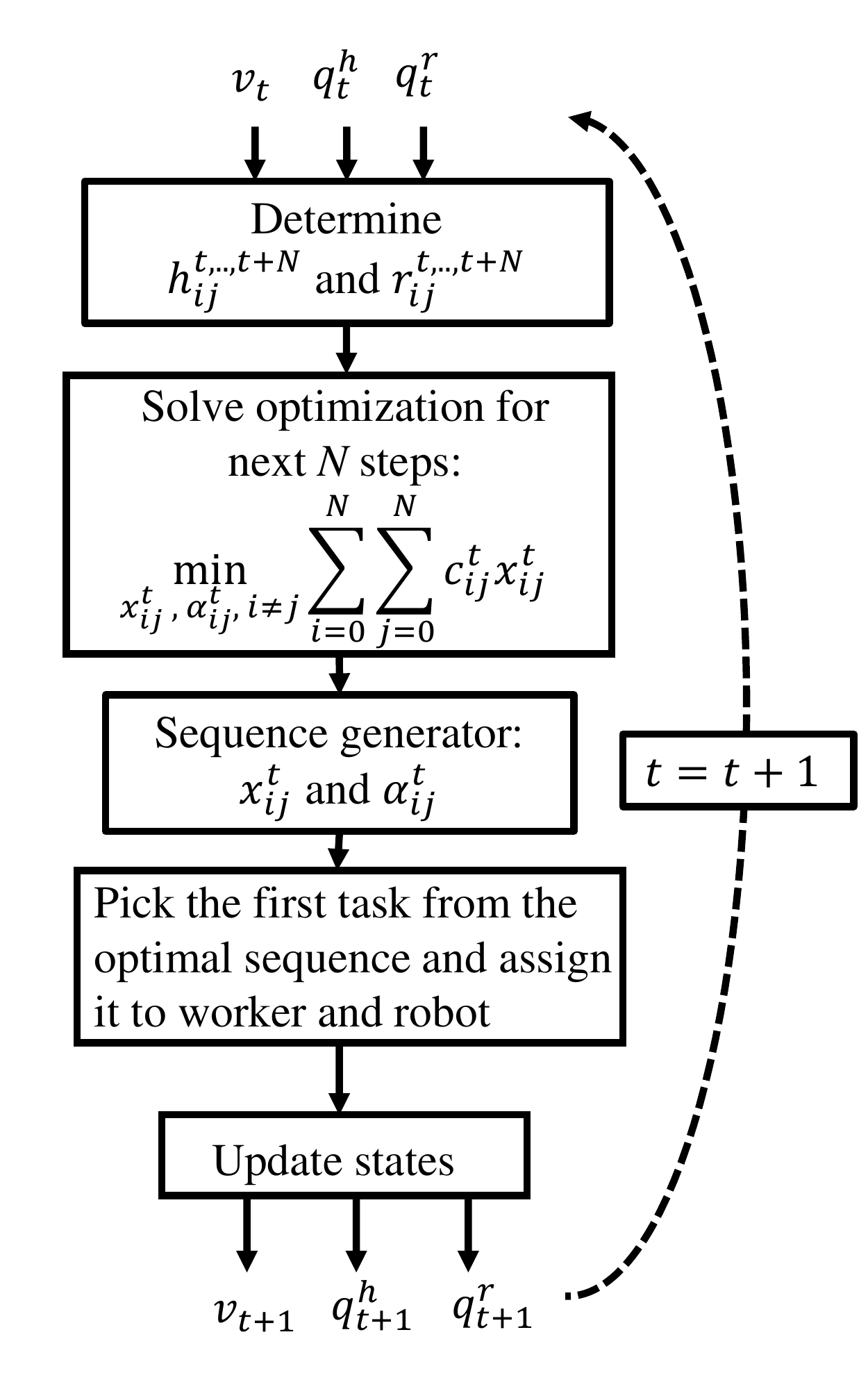}
	\caption{N-STEP DISASSEMBLY SEQUENCE PLANNER: SEQUENCE GENERATION AND OPTIMIZATION} 	\vspace{-15pt}
	\label{fig:control2}
\end{figure}
Since there is no actual dynamics in the task sequence planning, we exploit the concept of state-feedback control law so that a receding-horizon performance index, as shown in Fig.~\ref{fig:recede}, is minimized at each step using the following assertions. We assume that the ``states" of the system are the positions of the robot and the human operator and the set of remaining disassembly tasks. The ``performance index" explicitly considers the distances between disassembly tasks, robot and human operator and is changing based on real-time human motion. The ``constraints" of MPC correspond to the rules of precedence relations between the disassembly tasks and the safety conditions for human operation. After each step of completing a disassembly task, the task will be removed from the set of remaining tasks and the human may move to different locations. This yields new initial conditions for the next step. Hence, there is some resemblance between MPC and the proposed local optimal sequence planner in this paper.

By using the receding horizon technique, the optimization problem can be solved at each step without excessive computational cost. Also, real-time human motion can be explicitly included in the real-time receding horizon planner. The overview of such a receding horizon disassembly planner is illustrated in Fig. \ref{fig:control1} and its detail is shown in Fig. \ref{fig:control2}.

\vspace{-10pt}
\section*{VALIDATION}
In this section, the validation of the proposed disassembly sequence planner considering human and robot collaboration is presented. First, the toy box is placed arbitrarily on the table and a camera is used to identify all the to-be-disassembled components on the toy box using image processing and deep learning techniques (e.g., \cite{liang2019image}). 
\begin{figure}[!htbp] 
\vspace{-10pt}
	\centering
	\includegraphics[width=0.43\textwidth]{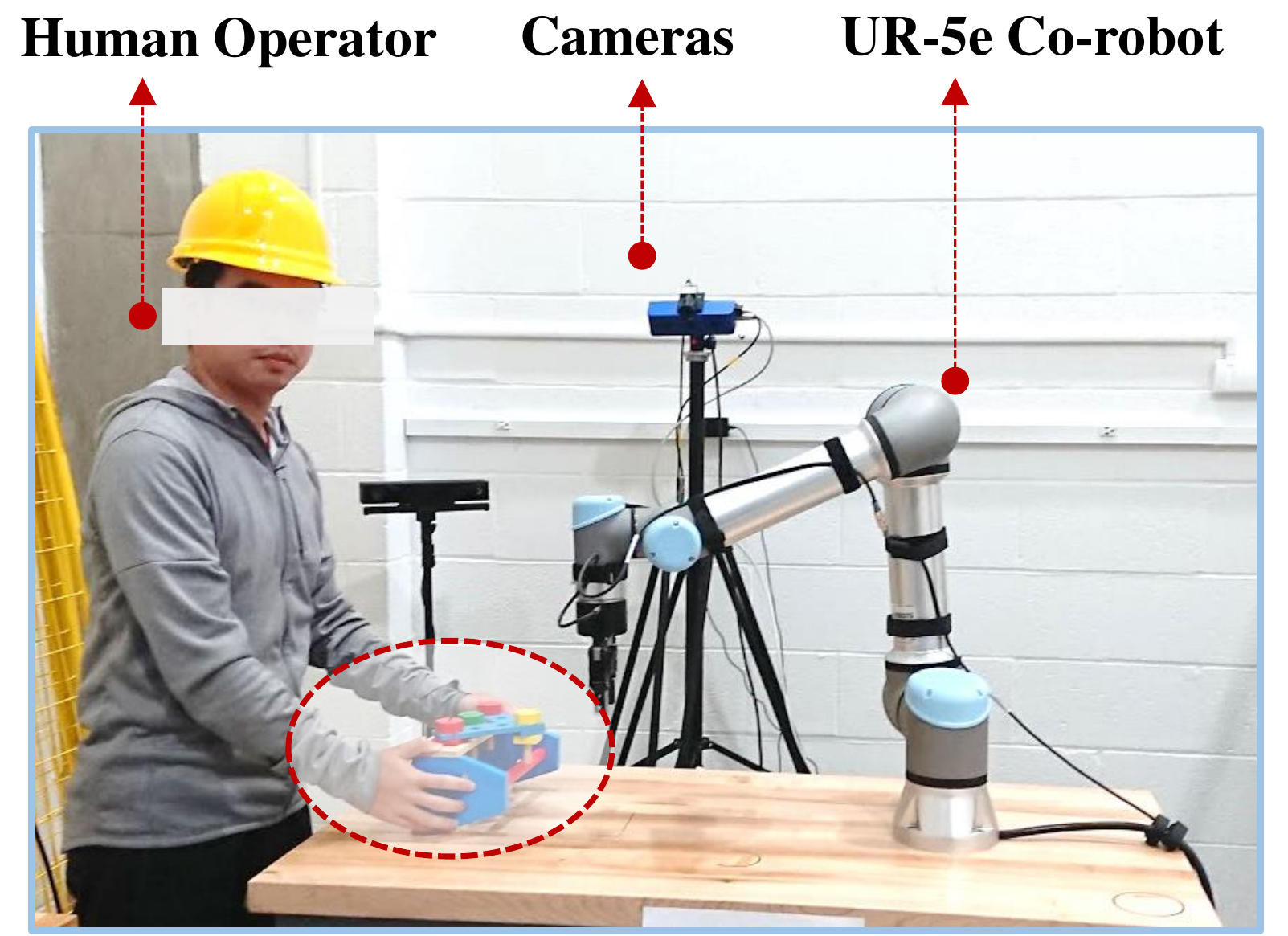}
	\caption{EXPERIMENTAL TEST SETUP}	\vspace{-15pt}
	\label{fig:exp_1}
\end{figure}
Second, the task planner optimizes the next $N$-step disassembly sequence over all feasible ones at round $t$. After getting the $N$-step local optimal sequence, one of the workers: the human operator or the robot, will be assigned to conduct the task. It is noteworthy that human operator in this experiment acts as an assistant when a disassembly task is performed by the robot. In these cases, the human operator will stay alongside the robot to complete the tasks collaboratively. And it is also assumed that all the screws are in reachable range of the human operator and the robot. More importantly, two variables of the toy box are arbitrarily given: the orientation and the component that contains hazard materials.

The experimental test that is used in this paper is as shown in Fig.~\ref{fig:exp_1}. The experimental setup consists of a universal robot (UR) that collaborates with the human operator and the requisite cameras used to identify the locations of the screws on the toy box, the robot, and the human operator. \vspace{-10pt}

\begin{figure}[!htbp]
	\centering
	\includegraphics[width=0.45\textwidth]{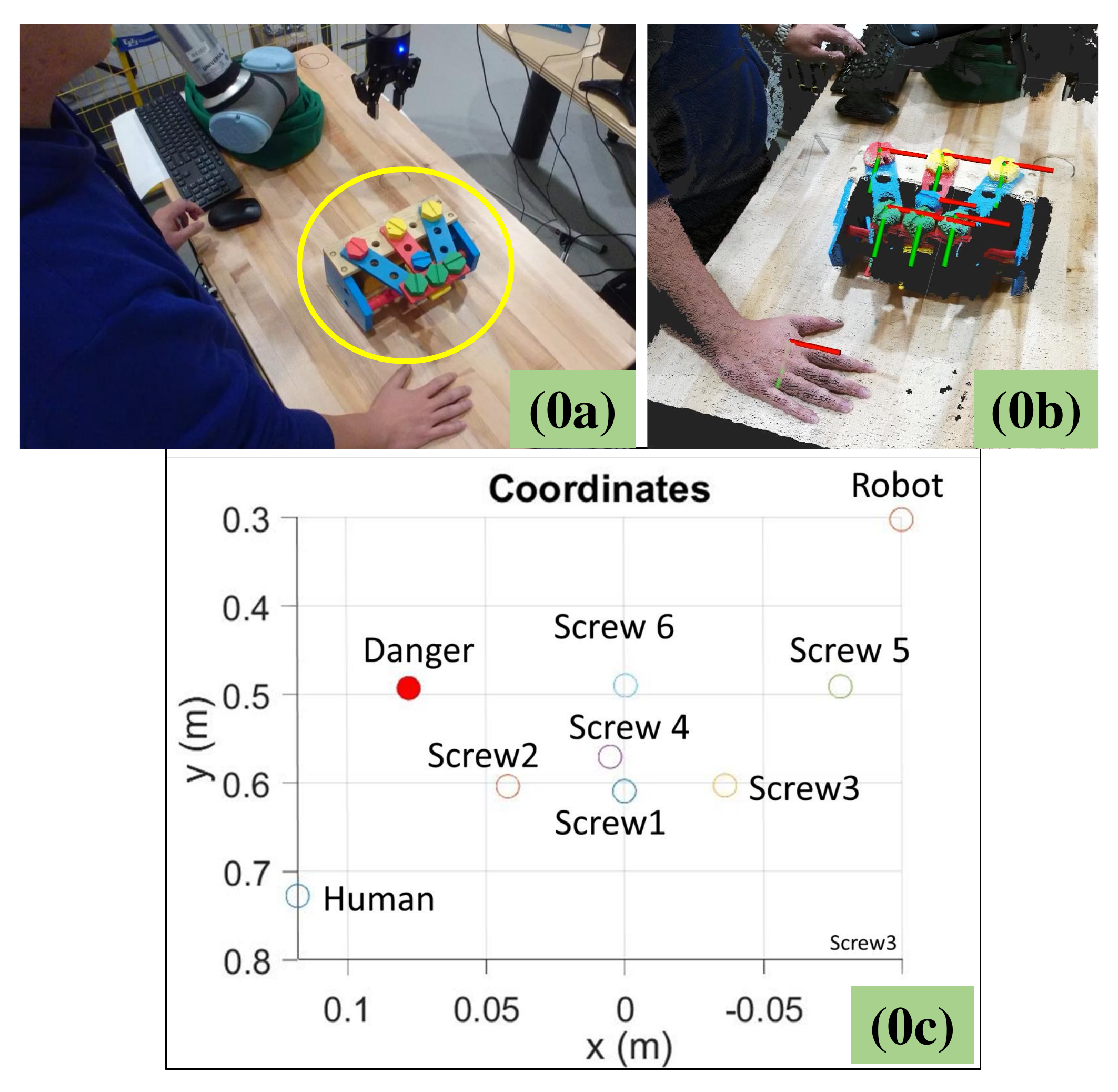}
	\caption{ROUND 0. IDENTIFICATION OF LOCATIONS OF SCREWS, ROBOT, AND HUMAN}
	\label{fig:exp5_0}
\end{figure}

\vspace{10pt}
\noindent
\textbf{Case study scenario: Screw 2 is closest to the robot and Screw 7 is unsafe for the human operator.}

\begin{figure}[b]
	\centering
	\includegraphics[width=0.42\textwidth]{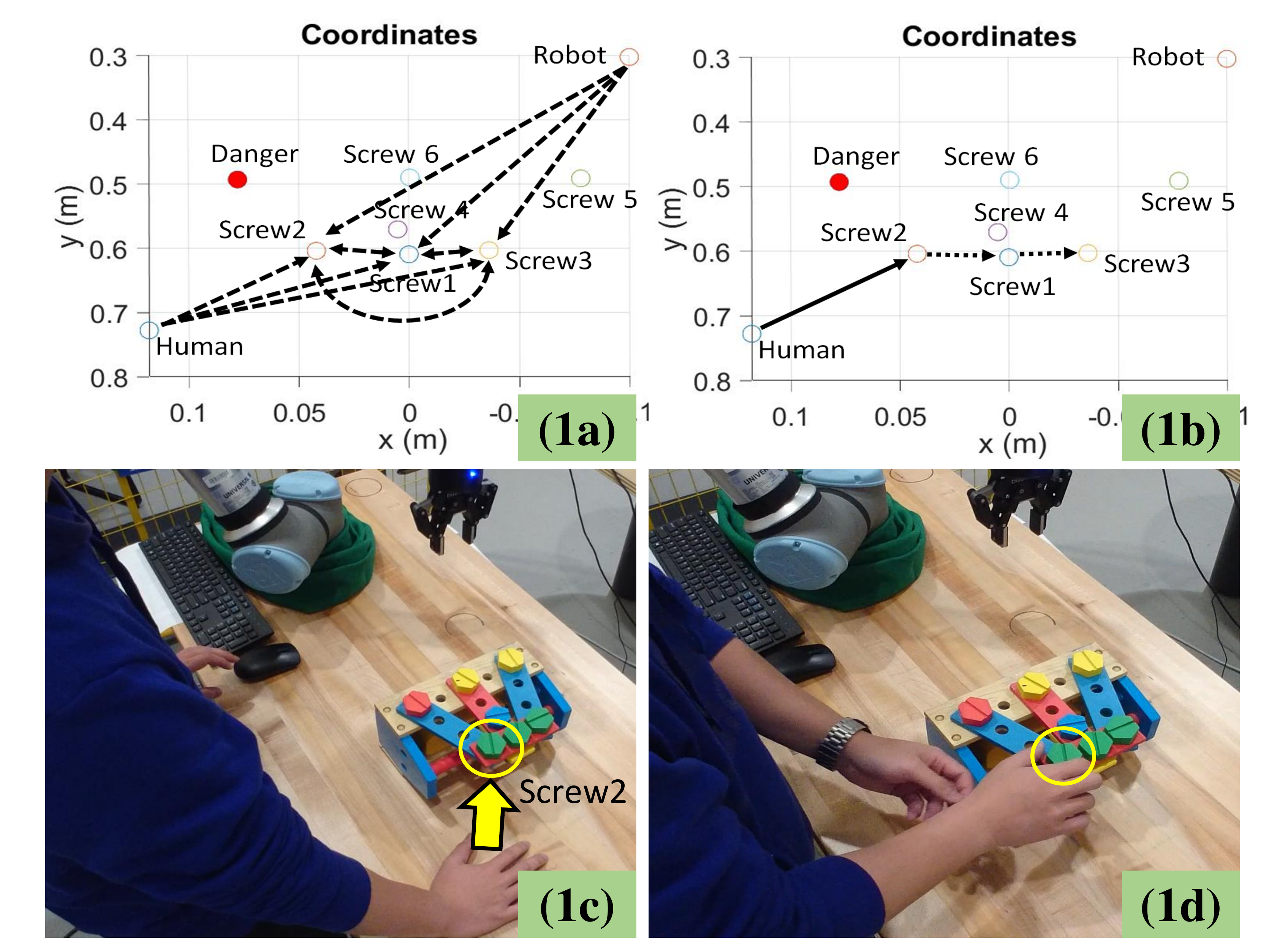}
	\caption{ROUND 1. HUMAN IS ASSIGNED TO SCREW 2}
	\label{fig:exp5_1}
\end{figure}

We firstly examine the scenario: the toy box is placed on the desk such that Screw 2 is the closest to the human operator, as shown in Fig.~\ref{fig:exp5_0}. In addition, the condition of a safety constraint at Screw 7 is emulated by using a red screw. The locations of the seven screws are identified by the camera using binary square fiducial markers \cite{aruco14} and then the information is sent to the computer for sequence planning. The planner then generates all possible disassembly sequences for the next three steps as depicted in Fig.~\ref{fig:exp5_1} (1a). Here, the hollow dots illustrate the positions of the robot and the human operator, and the screws that can be safely operated. The red dot denotes the screw that is unsafe for human operation. After receiving the information, the planner obtains the first three-step optimal sequence, in which the dashed arrows denote the feasible paths between the robot/human operator and the screws to be removed. Then, the solid arrow indicates the disassembly task of Screw 2 being assigned to the human operator as shown in Fig.~\ref{fig:exp5_1} (1b). Finally, the sequence planner ends the first round by commanding the human operator to disassemble Screw2, as demonstrated in Fig.~\ref{fig:exp5_1} (1c) and (1d). 

\begin{figure}[!htbp]
	\centering
	\includegraphics[width=0.42\textwidth]{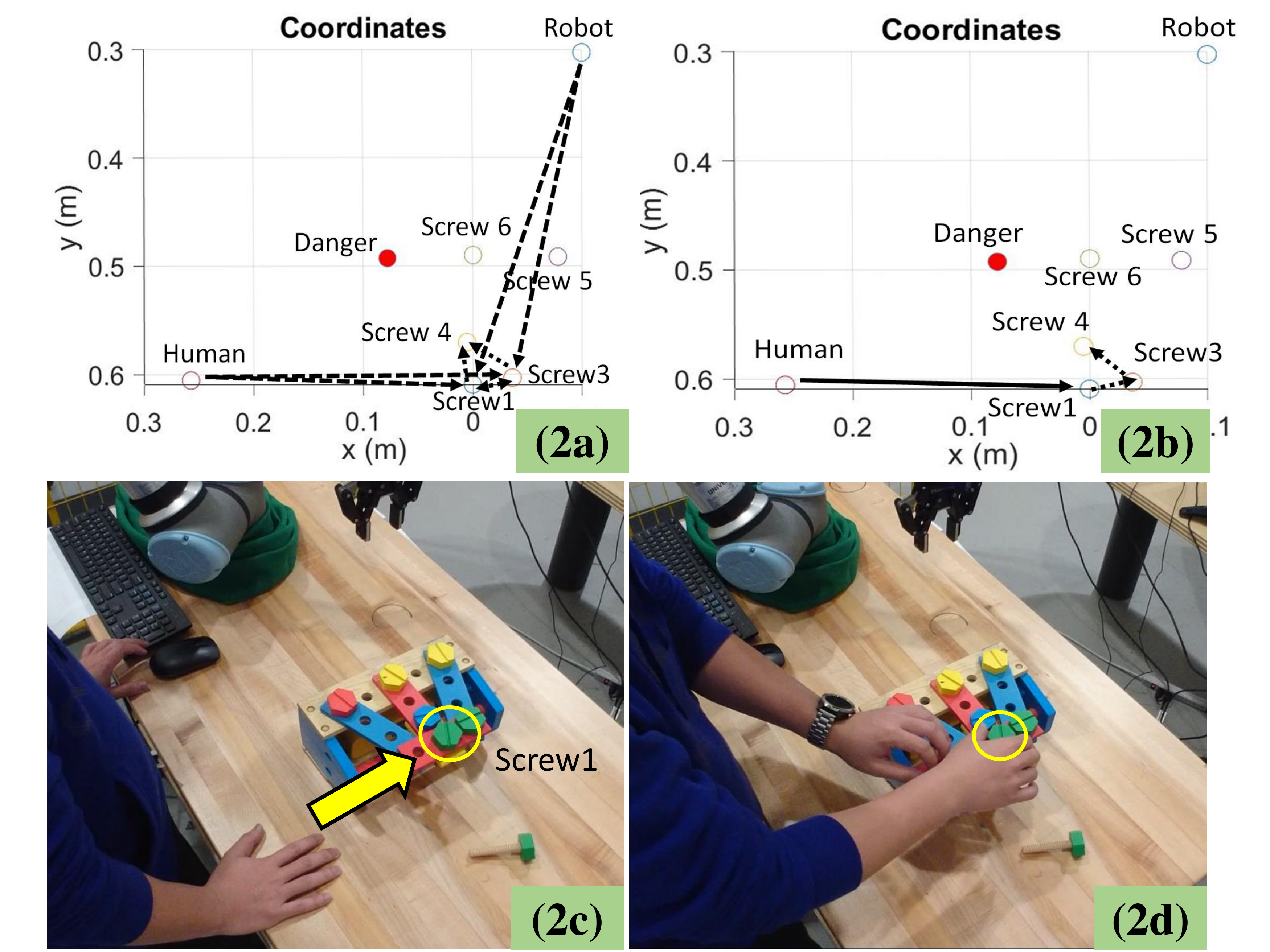}
	\caption{ROUND 2. HUMAN IS ASSIGNED TO SCREW 1}
	\label{fig:exp5_2}
\end{figure}

Similarly, in the second round, the sequence planner firstly generates feasible sequences for the next three steps in Fig.~\ref{fig:exp5_2} (2a), followed by the achieved local optimal sequence in Fig.~\ref{fig:exp5_2} (2b). And the next disassembly task is determined and the worker, either the robot or the human operator, is assigned, as illustrated in  Fig.~\ref{fig:exp5_2} (2b). Subsequently, the robot or the human operator starts moving from the initial position of round two. Then, the disassembly task is executed as shown in Fig.~\ref{fig:exp5_2} (2c) and (d). Because Screw 1 and Screw 3 have higher precedence relations than Screw 4, the task planner will always generate feasible sequences starting from Screw 1 and Screw 3.

\begin{figure}[!htbp]
	\centering
	\includegraphics[width=0.42\textwidth]{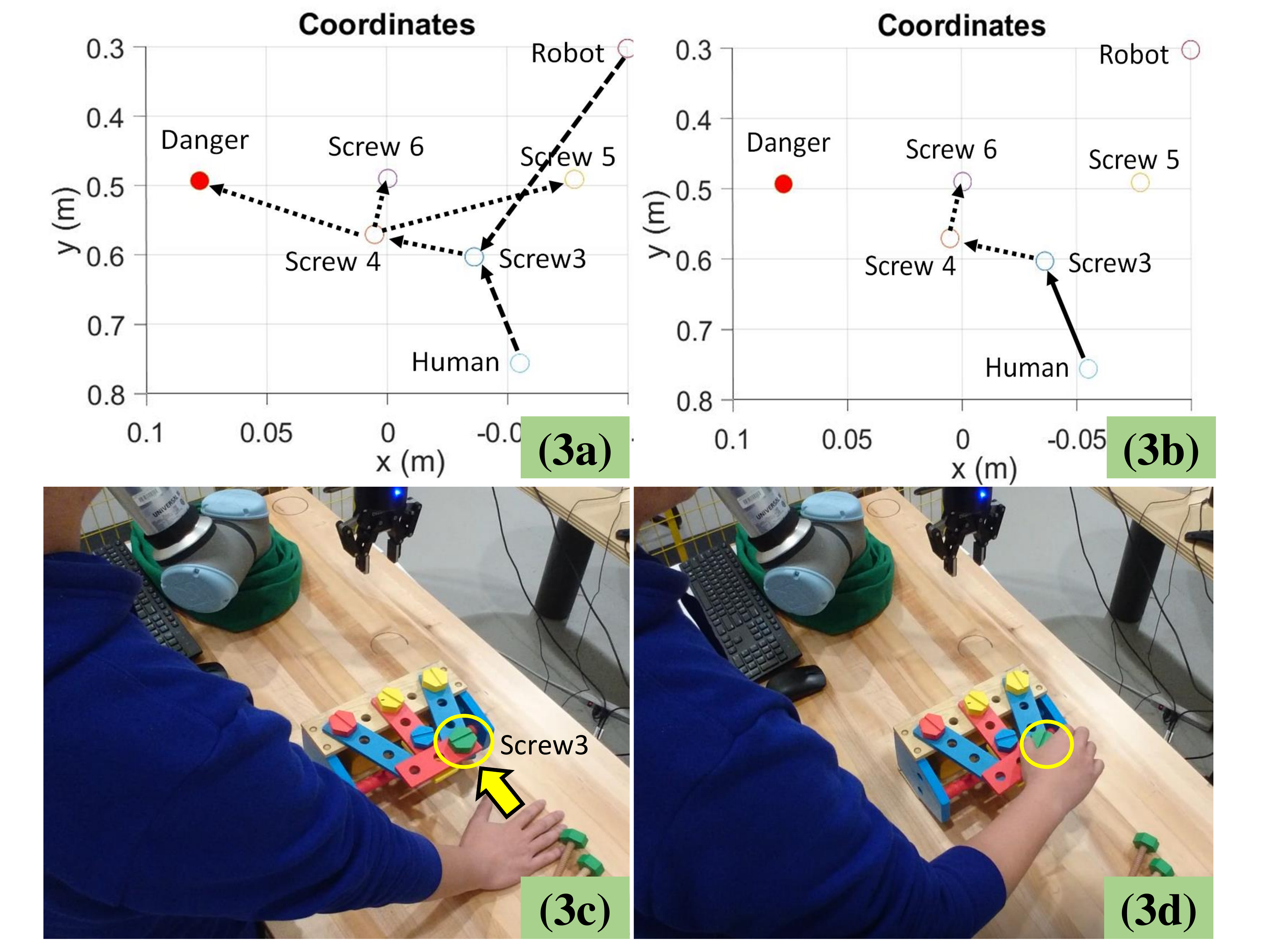}
	\caption{ROUND 3. HUMAN IS ASSIGNED TO SCREW 3}
	\label{fig:exp5_3}
\end{figure}

Fig.~\ref{fig:exp5_3} shows the process of determining the task and worker for the third disassembly task. The following local optimal sequence is found: Screw 3 being performed by the human operator, followed by Screw 4, and then Screw 6 disassembled by the human operator sequentially. Lastly, the sequence planner only picks the first task from the local optimal sequence and then executes with the human operator.

\begin{figure}[!htbp]
\vspace{-10pt}
	\centering
	\includegraphics[width=0.42\textwidth]{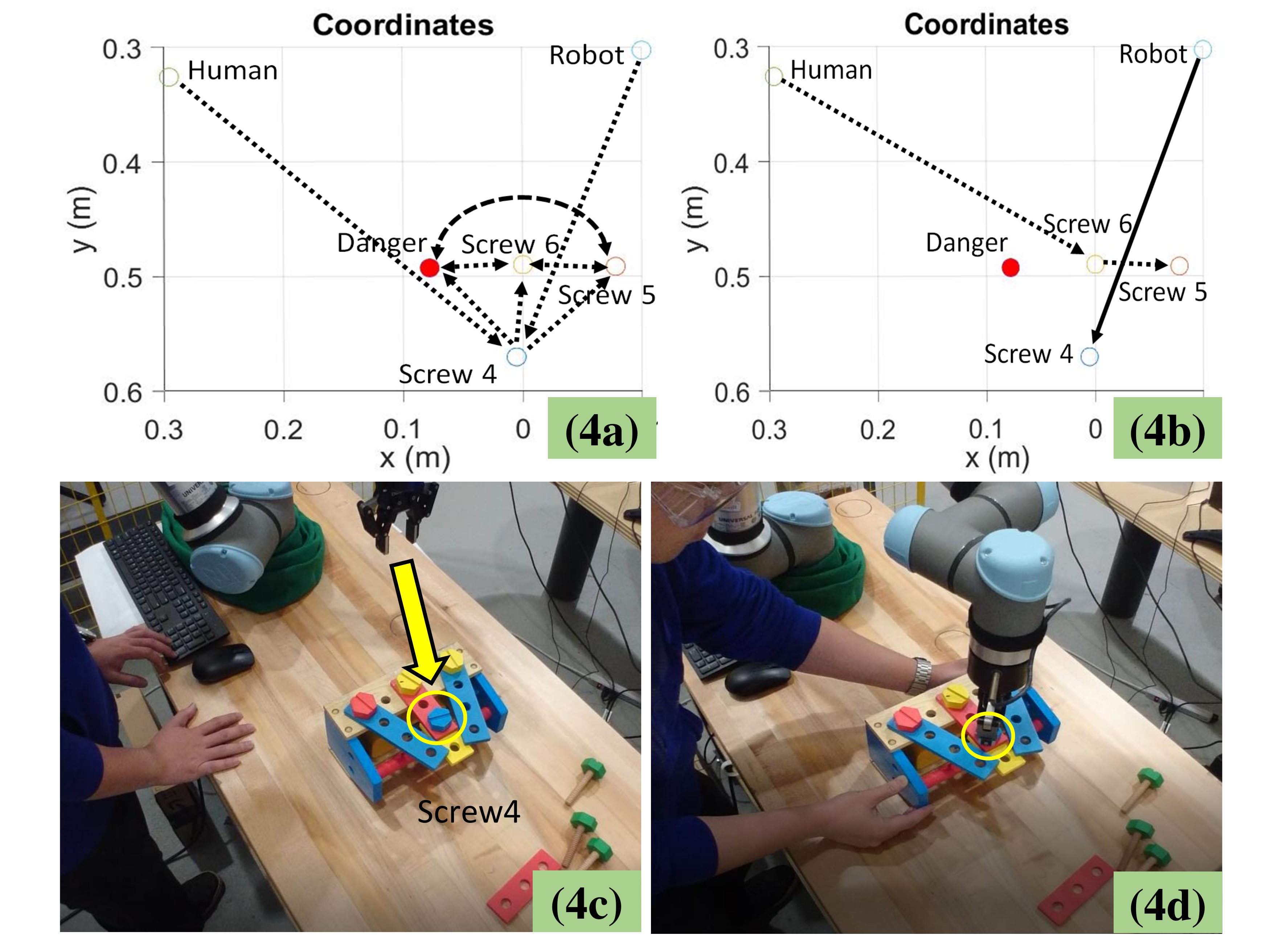}
	\caption{ROUND 4. ROBOT IS ASSIGNED TO SCREW 4}
	\label{fig:exp5_4}
\end{figure}

At round 4, the following local optimal sequence is found: Screw 4 to be removed by the robot, followed by Screw 6 being disassembled by the human operator, and then Screw 5 dismantled by human, as shown in Fig.~\ref{fig:exp5_4}. In this way, the first task in the local optimal sequence is picked and then executed by the robot. 

\begin{figure}[!htbp]
	\centering
	\includegraphics[width=0.42\textwidth]{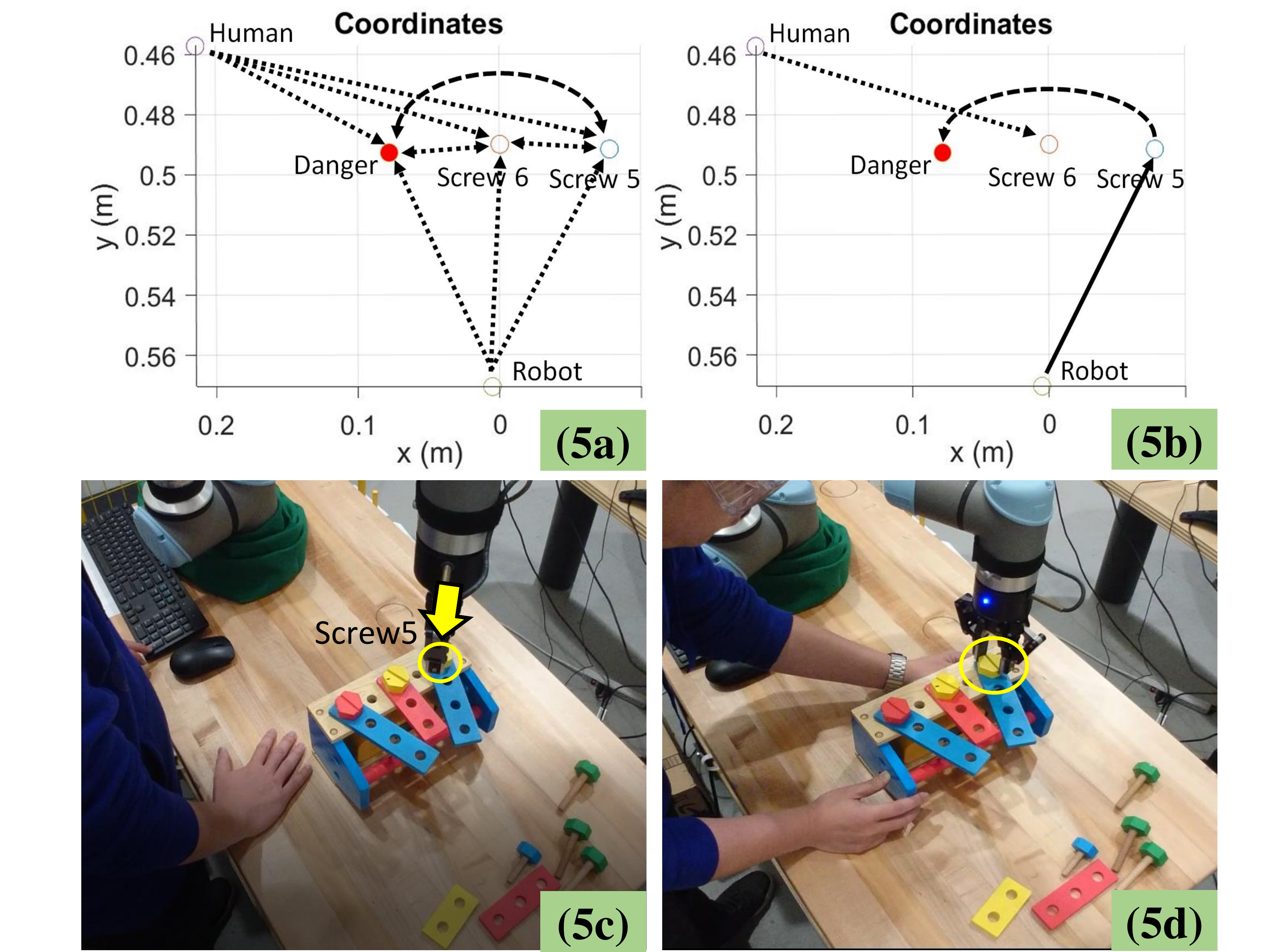}
	\caption{ROUND 5. ROBOT IS ASSIGNED TO SCREW 5}
	\label{fig:exp5_5}
\end{figure}

At round 5, the following local optimal sequence is achieved: Screw 5 being performed by the robot, followed by Screw 6 being disassembled by the human operator, and then Screw 7 being dismantled by the robot, as shown in Fig.~\ref{fig:exp5_5}. In this round, the first task in the local optimal sequence is chosen and then executed by the robot. \vspace{-5pt}

\begin{figure}[!htbp]
	\centering
	\includegraphics[width=0.42\textwidth]{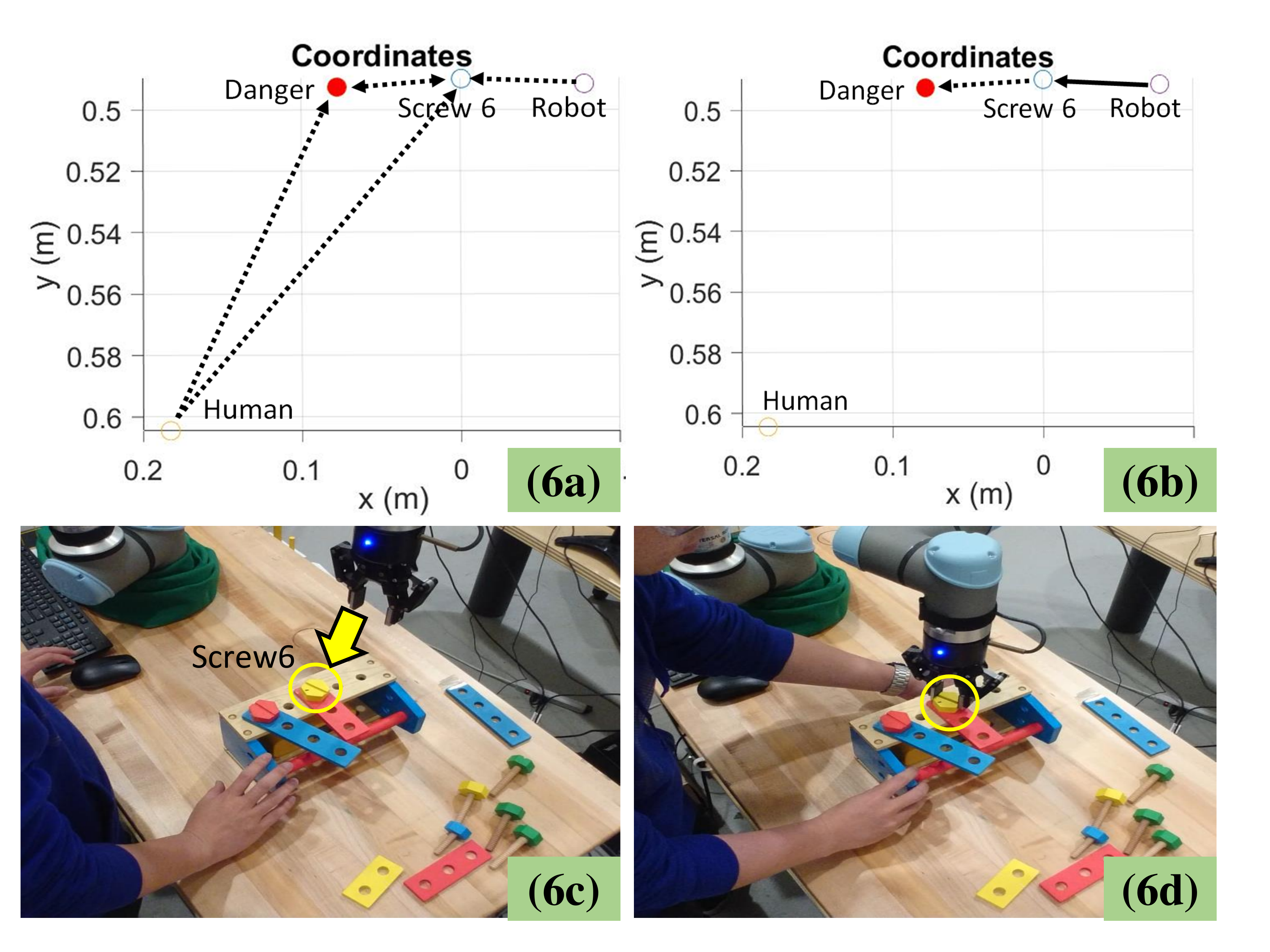}
	\caption{ROUND 6. ROBOT IS ASSIGNED TO SCREW 6}
	\label{fig:exp5_6}
\end{figure}

At round 6, because there are only two remaining tasks left, the following local optimal sequence will be obtained: Screw 6 being taken by robot, followed by Screw 7 being disassembled by robot as depicted in Fig.~\ref{fig:exp5_6}. And the first task in the local optimal sequence is picked and then performed by the robot. 

The last task is to disassemble the last screw that is assigned to the robot. In this experiment, Screw 7 is assumed to be unsafe for human operation. In consequence, the robot will be assigned to disassemble Screw 7 whether or not the disassembly cost of the robot is higher than the human operator. In other words, to ensure that the robot could handle the disassembly sequence safely, the decision variable $\alpha_{X7}$ in Fig.~\ref{fig:exp5_6}, is set to `1', to force the robot to execute the task, where X = 1,2,...7. 
\begin{figure}[!htbp]
	\vspace{-10pt}
	\centering
	\includegraphics[width=0.42\textwidth]{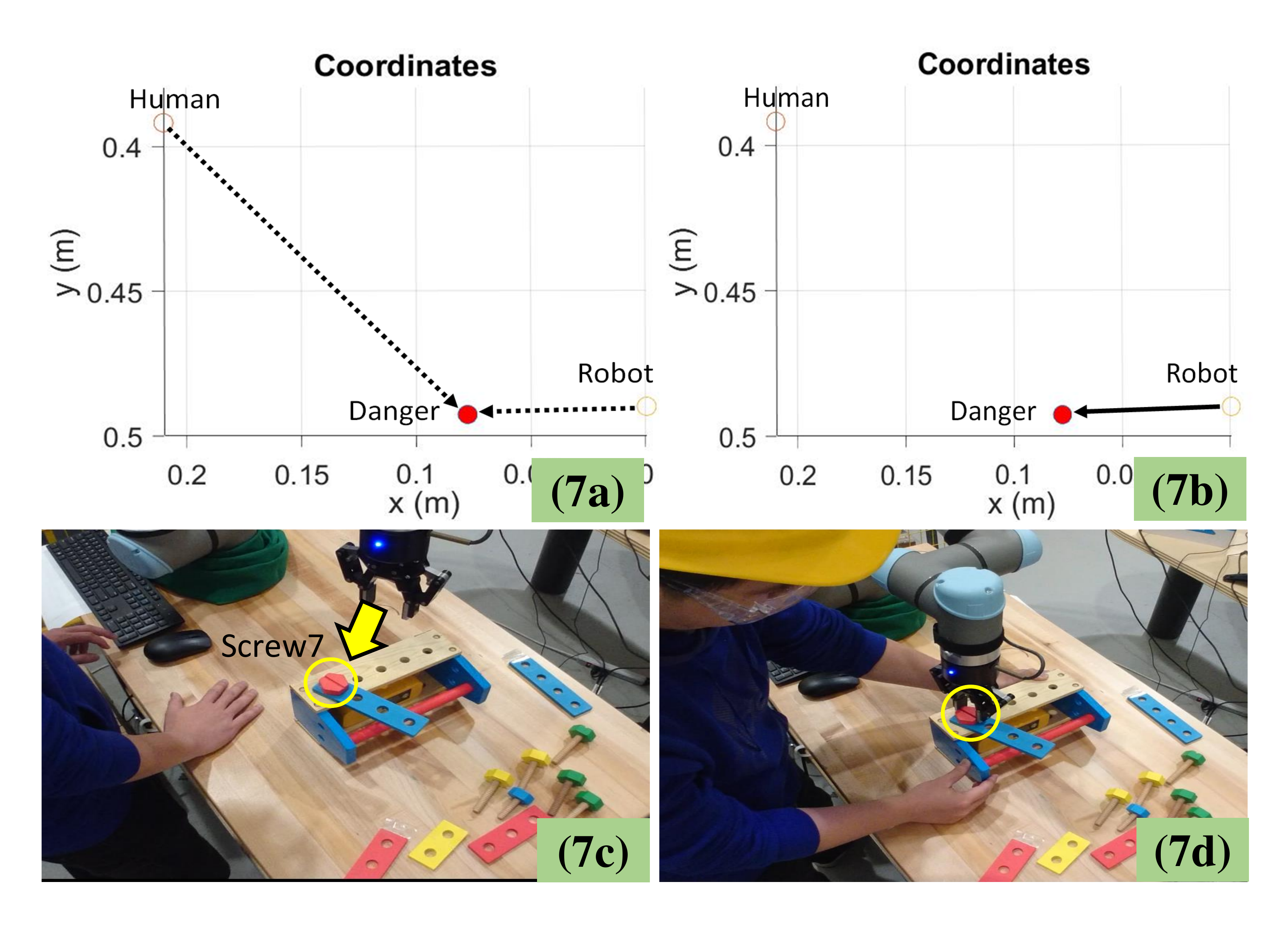}
	\caption{ROUND 7. ROBOT IS ASSIGNED TO SCREW 7}
	\label{fig:exp5_7}
	\vspace{-15pt}
\end{figure}

After the wooden toy box is placed on the table and the locations of the screws are identified at round 0, the whole process in the figures from  Fig.~{\ref{fig:exp5_1}} to Fig.~{\ref{fig:exp5_7}} can be concluded as follows, in which Steps (Xa) to (Xb) are processed in the computer and Steps (Xc) to (Xd) are executed by the human operator and the robot, where X = 1,2,...,7.\vspace{-5pt}
\begin{itemize}
	\item a. Possible three-step feasible sequences at each round are generated.
	\item b. Local optimal disassembly sequence for the next three-step tasks is achieved.
	\item c. The first task of the obtained sequence is selected and then the worker corresponding to the task is assigned.
	\item d. Human operator or the robot receives the command to execute the first task of the obtained optimal sequence.
	\item e. Human operator or the robot starts to work on the disassembly task. \vspace{-5pt}
\end{itemize}

\noindent
The experimental validation video is available via the following link: 
\url{http://zh.eng.buffalo.edu/PaperDemo/CalUBISFA2020ExpVideo.mp4}.

\vspace{-15pt}
\section*{CONCLUSIONS}
This paper presents the formulation of a disassembly sequence planning problem in human-robot collaboration setting into an optimization problem. Several important considerations have been taken into account. These considerations include different initial orientations of the to-be-disassembled product, varying disassembly cost with robot and human operator, feasible operations of the robot, and safety consideration for human operators. The optimization problem is solved in an receding horizon way and the real-time human motion is taken into consideration. A wooden toy box is used to emulate a real used-product and validate the proposed algorithm. The experimental test shows that the human operator and the robot complete the disassembly tasks collaboratively without violating the disassembly rules and the safety constraints.

\bibliographystyle{asmems4}

\vspace{-15pt}
\begin{acknowledgment}
This material is based upon work supported by the National Science Foundation - USA under Grant EFMA No. 1928595 and University at Buffalo Sustainable Manufacturing and Advanced Robotic Technologies (SMART) Fund. Any opinions, findings, and conclusions or recommendations expressed in this material are those of the authors and do not necessarily reflect the views of the National Science Foundation and the University at Buffalo.
\end{acknowledgment}
\vspace{-30pt}

\normalem
\bibliography{ref}
\end{document}